\documentclass[sigconf]{acmart}
\usepackage{xspace}

\newcommand{\dataset}{\textsf{CNER-UAV}\xspace}
\usepackage{array}
\usepackage{adjustbox}
\usepackage{url}
\usepackage{booktabs}
\usepackage{fontawesome5}
\usepackage{multirow}
\usepackage{enumitem}
\usepackage{hyperref}
\usepackage{url}

\AtBeginDocument{%
  \providecommand\BibTeX{{%
    \normalfont B\kern-0.5em{\scshape i\kern-0.25em b}\kern-0.8em\TeX}}}

\setcopyright{acmcopyright}
\copyrightyear{2018}
\acmYear{2018}
\acmDOI{XXXXXXX.XXXXXXX}

\copyrightyear{2024}
\acmYear{2024}
\setcopyright{acmlicensed}\acmConference[WWW '24 Companion]{Companion Proceedings of the ACM Web Conference 2024}{May 13--17, 2024}{Singapore, Singapore}
\acmBooktitle{Companion Proceedings of the ACM Web Conference 2024 (WWW '24 Companion), May 13--17, 2024, Singapore, Singapore}
\acmDOI{10.1145/3589335.3651446}
\acmISBN{979-8-4007-0172-6/24/05}


\begin{document}

\title{Can LLM Substitute Human Labeling?\\ A Case Study of Fine-grained Chinese Address Entity Recognition Dataset for UAV Delivery}


\author{Yuxuan Yao\textsuperscript{$\ast$}, Sichun Luo\textsuperscript{$\ast$}, Haohan Zhao\textsuperscript{$\ast$}, Guanzhi Deng, Linqi Song$^{\dagger}$}
\thanks{\textsuperscript{$\ast$}Equal Contribution\\$^{\dagger}$Coresponding Author}
\affiliation{%
  \institution{Department of Computer Science, City University of Hong Kong}
  \city{Hong Kong}
  \country{China}
}
\affiliation{%
  \institution{City University of Hong Kong Shenzhen Research Institute}
  \city{Shenzhen}
  \country{China}
}
\email{{yuxuanyao3-c, sichun.luo, haohazhao2-c, guanzdeng2-c}@my.cityu.edu.hk, linqi.song@cityu.edu.hk}

\begin{abstract}
We present \dataset, a fine-grained \textbf{C}hinese \textbf{N}ame \textbf{E}ntity \textbf{R}ecognition dataset specifically designed for the task of address resolution in \textbf{U}nmanned \textbf{A}erial \textbf{V}ehicle delivery systems. The dataset encompasses a diverse range of five categories, enabling comprehensive training and evaluation of NER models.
To construct this dataset, we sourced the data from a real-world UAV delivery system and conducted a rigorous data cleaning and desensitization process to ensure privacy and data integrity. The resulting dataset, consisting of around 12,000 annotated samples, underwent human experts and \textbf{L}arge \textbf{L}anguage \textbf{M}odel annotation. We evaluated classical NER models on our dataset and provided in-depth analysis.
The dataset and models are publicly available at \url{https://github.com/zhhvvv/CNER-UAV}.
\end{abstract}


\maketitle

\section{Introduction}
\label{sec:intro}

Named Entity Recognition (NER) categorizes entities in text based on predefined categories \cite{li2020survey}. Traditional grammar-based NER offers precision but requires significant effort from linguists. Statistical NER systems demand extensive manual annotation. Transformer models, like BERT \cite{devlin2018bert}, now dominate NER tasks \cite{vaswani2017attention,liu2022chinese}, showcasing remarkable performance.


The unmanned aerial vehicle (UAV) delivery system in China employs drones for goods delivery \cite{mt2021}. In this system, the address resolution module utilizes NER to convert raw user addresses into precise locations using language models \cite{liu2022chinese}. However, these models require specific fine-grained Chinese NER datasets for training, and there's a shortage of such datasets for Chinese, highly necessitating the creation of a new dataset.


In recent years, large language models (LLMs) have shown remarkable natural language understanding and generation abilities \cite{brown2020language,luo2024integrating,zhou2023solving,luo2023recranker,wang2023mathcoder}. While some studies have successfully used LLMs for labeling English datasets \cite{gilardi2023chatgpt,wang2021want}, research on their application in labeling fine-grained Chinese datasets is limited. This raises a natural question: \emph{Can LLMs surpass human performance in labeling fine-grained Chinese datasets with superior quality?} 

In this paper, we present \dataset, a fine-grained \underline{C}hinese \underline{NER} dataset for the \underline{UAV} delivery task. Specifically, \dataset~consists of five categories, ranging from buildings to rooms, allowing for fine-grained segmentation of Chinese addresses. It contains around 12,000 labeled samples. The raw data were collected from Meituan UAV delivery systems \cite{mt2021}, representing the actual data distribution. After carefully processing the raw data, we annotated the dataset using both \textbf{H}uman annotators and LLMs, including \textbf{G}PT-3.5~\cite{brown2020language} and ChatG\textbf{L}M~\cite{zeng2022glm}, resulting in 3 subsets: \dataset-\textbf{H}, \dataset-\textbf{G}, and \dataset-\textbf{L} correspondingly. We conducted experiments on all subsets to evaluate their performance with various baseline models. By comparing these datasets, we have determined that while LLMs can serve as supplementary annotation tools, they cannot entirely substitute manual annotation for fine-grained Chinese NER in the UAV delivery task.

In a nutshell, our contribution is threefold:
\begin{itemize}[leftmargin=*]
\item We introduce \dataset, a novel and fine-grained Chinese NER dataset specifically designed for the UAV delivery task. To the best of our knowledge, this is the most up-to-date and comprehensive address dataset from China for named entity recognition in UAV delivery services.
\item We compare the human annotation with the LLM annotation and conduct empirical studies demonstrating that humans outperform GPT-3.5 and ChatGLM in labeling the Chinese NER dataset, particularly for more fine-grained tags.
\item We evaluate several state-of-the-art baseline models and establish a benchmark for further research on this task. The evaluation results validate the usefulness of our proposed \dataset~dataset.
\end{itemize}

\section{Related Work}


\subsection{Chinese Named Entity Recognition Dataset}


Named Entity Recognition (NER) has been extensively studied in Natural Language Processing (NLP), and Chinese NER (CNER) datasets are pivotal for Chinese-specific NER models. The People's Daily (PD) corpus\footnote{\url{https://icl.pku.edu.cn/}} and larger-scale datasets like OntoNotes 4.0\footnote{\url{https://catalog.ldc.upenn.edu/LDC2011T13}} offer annotations, but challenges like domain coverage and noisy data remain. Datasets like WeiboNER \cite{peng2015named} address microblog NER, but the need persists for diverse, high-quality Chinese NER datasets catering to the language's uniqueness.

\subsection{Labeling Dataset with LLMs }

Labeling datasets with LLMs has gained recent attention for automating dataset creation and annotation, leading to cost savings and reduced effort \cite{wang2021want}. Collaborative approaches between humans and LLMs have also been explored, as seen in the WANLI dataset \cite{liu2022wanli}, and GPT-3.5's ability to outperform crowd-workers in text annotation tasks \cite{gilardi2023chatgpt}. While LLMs show promise in improving annotation quality and cost-efficiency, limited research explores their use in labeling fine-grained Chinese NER datasets. This study compensates for this research gap by focusing on LLMs' application in fine-grained Chinese address recognition dataset labeling.

\begin{figure}[htbp!]
    \centering
\includegraphics[width=0.49\textwidth]{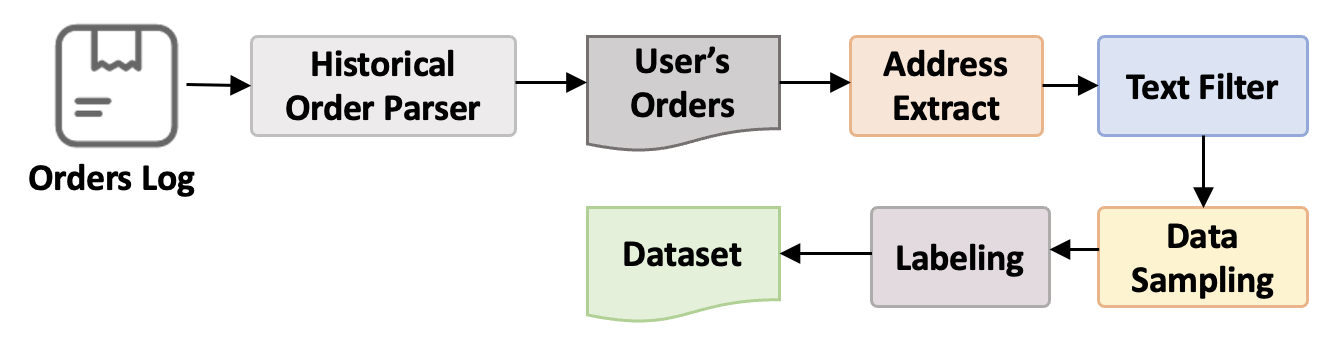}
    \caption{The pipeline of dataset construction.}
\label{fig:pipeline}

\vspace{-0.1in}
\end{figure}

\section{Dataset Construction}


\subsection{Overview}

Our dataset compilation process primarily adheres to the flowchart illustrated in Figure~\ref{fig:pipeline}. Initially, we parse the historical order logs to extract the user's order information. Subsequently, we identify and extract the addresses, applying predefined rules to filter out the relevant ones. Next, we proceed to sample and annotate the data, which involves labeling by human annotators, GPT-3.5 and ChatGLM. Further elaboration on the dataset collection and annotation procedures can be found in the subsequent sections.

\subsection{Dataset Collection}
The data originates from UAV delivery addresses provided by Meituan\footnote{\url{https://www.meituan.com/}}, which is one of the largest delivery service providers in China. When utilizing Meituan's online application, users can obtain services based on their personal preferences or practical address coverage through the following procedures: 1) Selecting a location point, which corresponds to the Point of Interest (POI) database, containing longitude and latitude coordinates. 2) Manually typing specific address details as auxiliary support. Our dataset is mainly sourced from users' textual descriptions in Shenzhen City and thus is a Chinese dataset, consisting of 12,000 pieces of data in each subset.

\subsection{Dataset Annotation}


To facilitate a more effective comparison between the LLMs and manual annotations, we established the following criteria for both manual and LLMs annotations:

1. Respect User Input and POI Information: Annotate all entities when multiple references point to a single location. 

2. Priority to Highest Probability Entity: When multiple entities of the same category appear in the address text, select the entity with the highest probability as the user's destination.

3. Filter Ambiguous or Unlocatable Addresses: Exclude addresses with ambiguous descriptions or those that cannot be geolocated based on longitude and latitude.






Specifically, the annotation prompt template for GPT-3.5 and ChatGLM is designed as:\\
\texttt{Extract information from the address by referring to the examples, extract POI, building, unit, level, and room information, output them in JSON format.\\
Example: \{examples\}\\
Instruction: \{instructions\}}\\
\begin{table}[h]
\centering
\caption{Count of sentences and tags in the dataset.}
\vspace{-0.05in}
\label{dataset compilation}
\begin{adjustbox}{width=\linewidth}
\begin{tabular}{lllll}
\toprule
Dataset   & Sentence Count & Tag Count(GPT) & Tag Count (Manual) & Tag Count(GLM)\\ \midrule
Train & 10150 & 58778 & 53220 & 73566\\ 
Test  & 1269 & 7396 & 6731 & 9211\\
Validation& 1269 & 7466 & 6690 & 9294\\
Overall & 12688 & 73640 & 66641 & 73566\\

\bottomrule
\end{tabular}
\end{adjustbox}
\end{table}
\begin{table}[h]
\fontsize{7}{7.5} \selectfont
\centering
\caption{CNER-UAV Dataset Description.}
\vspace{-0.05in}
\label{dataset description}

\scalebox{1}{
\begin{tabular}{lllll}
\toprule
 &Tags   & Train & Test & Validation \\ \midrule
\multirow{5}{*}{CNER-UAV-L}   
&Building & 11403 & 1441 & 1422\\
&Unit   & 6542 & 816 & 828\\
&Level   & 3489 & 447 & 441\\
&Room   & 3209 & 422 & 413\\
&Others  &108324 &13585 & 13962\\ \midrule

\multirow{5}{*}{CNER-UAV-G}   
&Building & 7513  & 960 & 972\\
&Unit   & 1484 & 215 & 190\\
&Level   & 8155 & 1019 & 1019\\
&Room   & 8280 & 1018 & 1029\\
&Others  &124153 &15538 & 15869\\ \midrule

\multirow{5}{*}{CNER-UAV-H}   
&Building & 8125 & 1037 & 1045\\ 
&Unit   & 1367 & 184 & 170\\
&Level   & 8642 & 1079 & 1070\\
&Room   & 7522 & 928 & 931\\
&Others & 140635 & 17560 & 18003\\
\bottomrule
\end{tabular}
}
\end{table}
\begin{table*}[h]
\centering
\caption{Sample Tagged Sentences, where ‘Others' is represented as ‘O', ‘Building' as 'B', ‘Level' as ‘L' and ‘Room' as ‘R'}
\label{dataset examples}
\begin{adjustbox}{width=\linewidth}
\begin{tabular}{c c c c c}
\toprule
Sentence   & English Translation & Tag \\ \midrule

\includegraphics[scale=0.55]{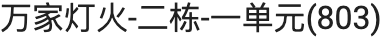} 
 & Wanjia Lighting - Building 2 - Unit 1 (803) & O O O O O B-B I-B O B-U I-U I-U O B-L B-R I-R O\\

\includegraphics[scale=0.55]{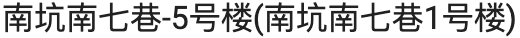} 
& Building 5, Nanqi Lane, Nankeng (Building 1, Nanqi Lane, Nankeng) & O O O O O O B-B I-B I-B O O O O O O B-B I-B I-B O \\
\includegraphics[scale=0.55]{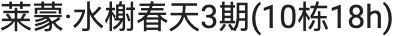}
& Lemon·Shuixie Spring Phase 3 (Building 10 18h) & O O O O O O O O O O B-B I-B B-L I-L B-R O\\

\bottomrule
\end{tabular}
    \end{adjustbox}
\end{table*}
\noindent where examples are annotated addresses based on the BOI standard, and instructions encompass Chinese language conventions and practical application scenarios, serving as both restrictions and suggestions. 

Furthermore, considering Chinese language patterns, we devised regularization formulas to correlate the annotated addresses with Beginning of Inside (BOI) \cite{lample2016neural} labels, forming the annotated dataset.

\subsection{Dataset Statistics}
Table~\ref{dataset compilation} presents an overview of sentences and tags within our datasets. Table~\ref{dataset description} shows the statistics of different labels in 3 subsets. Buildings are labeled as ‘Building', units as ‘Unit', floors as ‘Level', rooms as ‘Room', and non-entities are denoted as ‘Others'.
We could observe from the table, in the 'Building' classification, GPT annotations are approximately 7.47\% less compared to manual labeling. Conversely, in the 'Unit' category, GPT's annotations exceeded human labeling by about 9.76\%. These disparities suggest a potential oversight in specific 'Unit' labels by GPT, possibly arising from a deficiency in relevant knowledge or an inherent limitation within the model.

For more clarity, some example sentences with tagged entities are mentioned in Table~\ref{dataset examples}.

\begin{table*}[]
\fontsize{7}{8} \selectfont
\centering
\caption{F1 score and accuracy of various transformer and normal models for BOI notation using the UAV-NER dataset, where Acc means accuracy, G indicates dataset annotated by GPT-3.5 and H represents manually labeled dataset.}
\bgroup
\def\arraystretch{1.2}
\label{tab:main_result}
\begin{tabular}{llllllllllllllll}
\toprule
\multicolumn{2}{c}{\multirow{2}{*}{}} & \multicolumn{2}{c}{BERT}        & \multicolumn{2}{c}{ELECTRA}     & \multicolumn{2}{c}{XLNET}       & \multicolumn{2}{c}{RoBERTa}     & \multicolumn{2}{c}{ERNIE}       & \multicolumn{2}{c}{LERT}        & \multicolumn{2}{c}{PERT}        \\ \cmidrule(lr){3-4}\cmidrule(lr){5-6}\cmidrule(lr){7-8}\cmidrule(lr){9-10}\cmidrule(lr){11-12}\cmidrule(lr){13-14}\cmidrule(lr){15-16}
\multicolumn{2}{l}{}      & Acc $\uparrow$  & F1 $\uparrow$         & Acc $\uparrow$  & F1 $\uparrow$    & Acc $\uparrow$  & F1 $\uparrow$    & Acc $\uparrow$   & F1 $\uparrow$    & Acc     $\uparrow$    & F1  $\uparrow$   & Acc $\uparrow$   & F1      $\uparrow$    & Acc $\uparrow$   & F1  $\uparrow$          \\ \midrule
\multirow{3}{*}{Building}     & L & 70.69& 82.73& 66.98& 80.15& 69.82& 82.13& 70.76& 82.77& 70.01&
        82.77& 70.92& 82.89& 72.42& 83.86 \\
& G     & 95.19          & 96.13          & \textbf{95.04}          & 96.02          & \textbf{94.88}          & 95.87          & \textbf{95.74}          & 96.59          & \textbf{95.19}          & 96.14          & 95.67          & 96.53    & \textbf{96.38}          & 97.11          \\
                              & H     & \textbf{95.74} & \textbf{96.89} & 94.88 & \textbf{96.23} & 94.48 & \textbf{95.95} & 95.67 & \textbf{96.83} & 94.96 & \textbf{96.29} & \textbf{96.30} & \textbf{97.30} & 96.22 & \textbf{97.25} \\ \midrule
\multirow{3}{*}{Unit}         & L & 60.20 & 63.70 & 57.13& 61.42& 61.78& 65.48& 59.02& 63.33& 59.03& 63.41& 61.39& 64.75& 61.31& 65.25 \\
& G     & 94.09          & 76.64          & 94.01          & 76.97          & 94.01          & 76.54          & 93.38          & 74.85          & 93.70          & 74.85          & 94.09          & 77.48          & 94.41          & 78.02          \\
                              & H     & \textbf{98.27} & \textbf{92.99} & \textbf{98.11} & \textbf{92.26} & \textbf{98.11} & \textbf{92.26} & \textbf{98.50} & \textbf{93.97} & \textbf{97.95} & \textbf{91.72} & \textbf{98.42} & \textbf{93.63} & \textbf{98.66} & \textbf{94.60} \\ \midrule
\multirow{3}{*}{Level}        & L &71.32& 51.34& 67.77& 46.68& 69.98& 51.34& 68.95& 50.38& 68.88&
        50.31& 70.45& 51.98& 72.34& 53.14 \\

& G     & 91.10          & 94.45          & 90.86          & 94.35          & 90.54          & 94.13          & 90.94          & 94.35          & 90.94          & 94.35          & 89.83          & 93.65          & 90.31          & 93.97          \\
                              & H     & \textbf{96.53} & \textbf{97.93} & \textbf{95.82} & \textbf{97.50} & \textbf{95.51} & \textbf{97.31} & \textbf{95.82} & \textbf{97.51} & \textbf{96.22} & \textbf{97.74} & \textbf{96.30} & \textbf{97.79} & \textbf{97.08} & \textbf{98.27} \\ \midrule
\multirow{3}{*}{Room}         &L & 68.64& 41.82& 65.41& 35.35& 67.38& 41.85& 66.75& 41.55& 66.76& 41.58& 68.24& 42.35& 69.74& 41.82\\

& G     & 79.91          & 86.56          & 78.88          & 86.14          & 79.67          & 86.78          & 79.59          & 86.49          & 79.04          & 86.49          & 79.28          & 86.21          & 79.20          & 86.22          \\
                              & H     & \textbf{94.96} & \textbf{96.52} & \textbf{92.99} & \textbf{95.14} & \textbf{92.51} & \textbf{94.82} & \textbf{94.48} & \textbf{96.18} & \textbf{94.48} & \textbf{96.19} & \textbf{94.33} & \textbf{96.08} & \textbf{95.90} & \textbf{97.18} \\ \midrule
\multirow{3}{*}{Others}       &L &54.14& 70.21& 51.77& 68.22& 53.59& 69.78& 52.17& 68.57& 52.21&
        68.6 & 53.27& 69.51& 54.85& 70.81\\

& G     & 78.17          & 87.75          & 78.88          & 88.19          & 77.86          & 87.05          & 79.35          & 88.49          & 79.12          & 88.49          & 78.17          & 87.75          & 79.35          & 88.49          \\
                              & H     & \textbf{91.80} & \textbf{95.73} & \textbf{90.62} & \textbf{95.08} & \textbf{90.15} & \textbf{94.82} & \textbf{91.57} & \textbf{95.60} & \textbf{91.10} & \textbf{95.34} & \textbf{92.28} & \textbf{95.98} & \textbf{93.14} & \textbf{96.45} \\ \midrule
\multirow{3}{*}{Overall @macro}  & L &65.00& 61.96& 61.81& 58.36& 64.51& 62.12& 63.53& 61.32& 63.38& 61.33& 64.85& 62.3& 66.13& 62.98\\

& G  & 87.69 & 88.31 & 87.53 & 88.33 & 87.39 & 88.07 & 87.80 & 88.15 & 87.60 & 88.06 & 87.41 & 88.32 & 87.93 & 88.76 \\
& H & \textbf{95.46} & \textbf{96.01} & \textbf{94.48} & \textbf{95.24} & \textbf{94.15} & \textbf{95.03} & \textbf{95.21} & \textbf{96.02} & \textbf{94.94} & \textbf{95.46} & \textbf{95.53} & \textbf{96.16} & \textbf{96.23} & \textbf{96.75}

\\\bottomrule

\end{tabular}
\egroup
\vspace{-1em}
\end{table*}

\section{Experiment}
\subsection{Experiment Setup}
\textbf{Baselines} \quad Transformer-based models have found success across multiple NLP tasks, including named entity recognition, eliminating the need for manual feature engineering. 
To evaluate the dataset, we implement the following baselines: BERT \cite{devlin2018bert}, ELECTRA \cite{clark2020electra}, RoBERTa \cite{liu2019roberta}, XLNET \cite{yang2019xlnet},
ERNIE \cite{zhang2019ernie},LERT \cite{cui2022lert}, and PERT \cite{cui2022pert}.
 \smallskip 

\noindent\textbf{Evaluation Metrics} \quad In line with \cite{litake2022l3cube}, we evaluated various NER models using standard performance metrics, including Accuracy and F1-score. We assessed the performance at both the class level and overall performance.

\subsection{Results and Discussion}
The experimental results are presented in Table~\ref{tab:main_result}. The table reveals a notable disparity in annotation effectiveness, with ChatGLM-6B performing significantly worse compared to other annotation methods. In essence, it falls short of meeting the practical requirements of the UAV system. Our analysis suggests that this performance discrepancy can be attributed to ChatGLM's primary training data, which mainly comprises Chinese conversations. Without fine-tuning specifically for NER annotation tasks, its applicability to production needs remains inadequate. 


We've noticed that GPT annotation performs competitively with manual annotation when labeling “Buildings." However, it significantly lags behind in accurately labeling “Rooms" and “Others" compared to human annotation. This underscores the limitations of GPT annotation and highlights the challenges GPT models encounter in achieving fine-grained understanding of Chinese text. A comprehensive dataset review reveals several key insights. Building information is relatively straightforward and often presented in formats like “No. 2 Building." However, variations abound in the descriptions of units or levels due to diverse developer naming preferences and users' desire for food delivery convenience. Users may forego conventional terms like “3rd floor" in favor of more familiar references such as “the floor where Star Bookstore is located." Consequently, GPT-3.5's annotation effectiveness noticeably diminishes when handling such diverse data.

Furthermore, it is worth noting that all language models achieve over 90\% accuracy in each category of the human-annotated dataset. This result demonstrates the effectiveness and utility of our proposed dataset, affirming its value for UAV delivery task.

\begin{figure*}[htbp!]
    \centering
    
\includegraphics[width=0.8\textwidth]{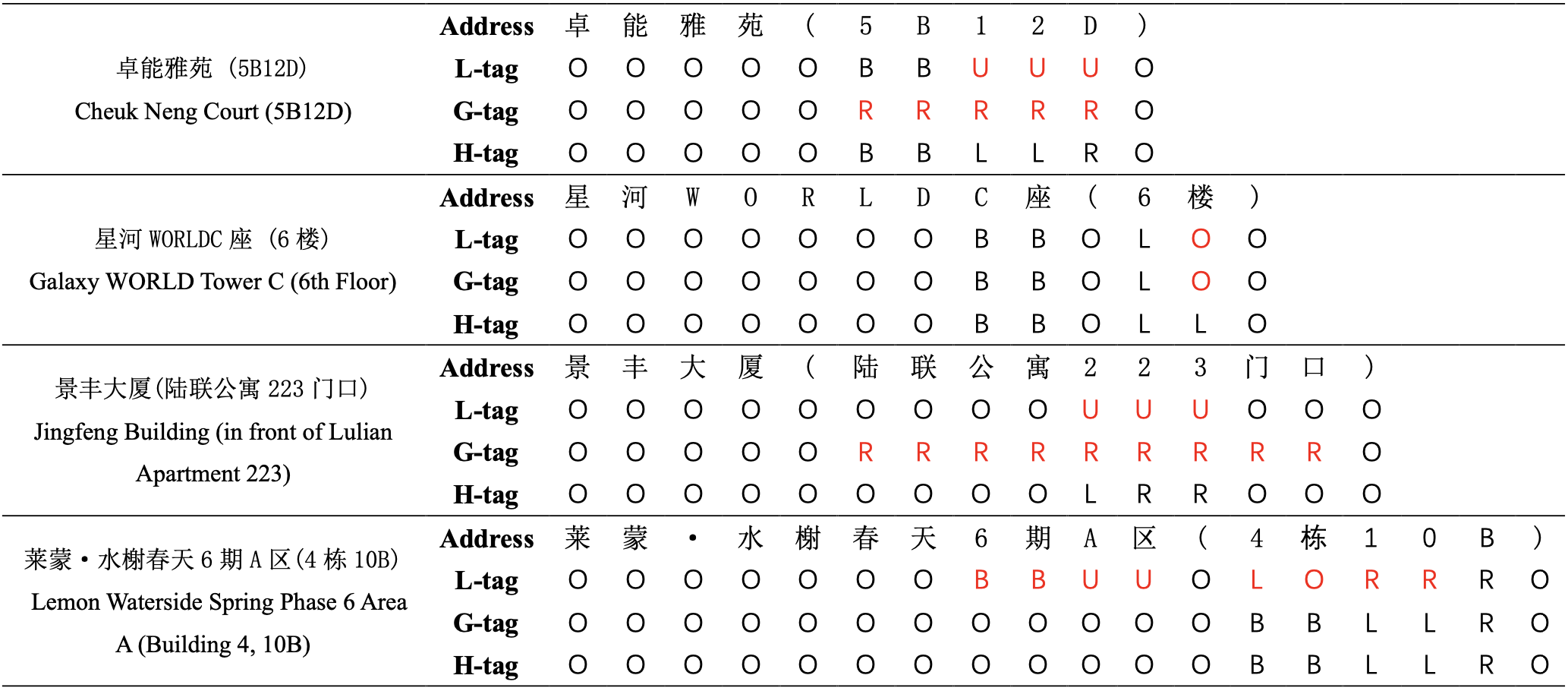}
    \caption{Bad Cases of CNER-UAV-G and CNER-UAV-L datasets, where labels from CNER-UAV-G dataset are shown as G-tag, CNER-UAV-L as L-tag,CNER-UAV-H as H-tag}
\label{fig:badcase}
\end{figure*}

\subsection{Case Study}
The experimental findings suggest that currently, large language models cannot fully substitute manual annotation. Figure~\ref{fig:badcase} lists four distinct types of bad cases in language models' address tagging:

1. Label Ambiguity:
In cases where users input a combination of building and floor details, for instance, “Cheuk Neng Court(5B12D)", GPT-3.5 cannot extract the ‘5B' as B-tag, ‘12' as L-tag, and ‘D' as R-tag, while ChatGLM cannot separate the floor and room labels.

2. Incomplete Entity Labeling:
LLMs sometimes provide incomplete entity labeling. For example, in the input ‘6th floor,' GPT-3.5 and ChatGLM only label ‘6th' and overlook ‘floor' in Chinese.

3. Consecutive Number Confusion:
When consecutive numbers are entered, LLMs face challenges in correctly distinguishing between floor and room numbers. For instance, they may fail to recognize the first “2" in “223" as a “Level" tag. In contrast, human annotation effortlessly avoids such issues.

4. Incapacity for Long Text:
Compared to GPT-3.5, ChatGLM exhibits a significant decline in its ability to parse Chinese long-text addresses, exhibiting severe label confusion. This aligns with our experimental findings.

To summarize, LLMs excel with singular and standardized address information but fall short in handling semantically rich and syntactically flexible addresses. Enhancing LLM's Chinese comprehension capabilities is a promising research avenue.


\section{Ethics Information }

The \dataset dataset was collected in the first half of 2023. The task of data annotation was outsourced to an external company. The data collection process was approved by the Meituan Ethics Review Board, ensuring that the process adhered to ethical guidelines.
Meituan is responsible for data desensitization and the deletion of user-sensitive information, ensuring that the privacy and confidentiality of the users are well-protected.


\section{Conclusion}
We present a fine-grained Chinese NER Dataset for the UAV delivery system, referred to as \dataset. The dataset consists of three subsets, but for practical applications, we are releasing two of them, one labeled by human annotators and the other labeled by GPT-3.5. These subsets are derived from real-world system logs, reliably reflecting user behavior. We perform experiments using popular baseline methods on both subsets. Empirical results demonstrate that human annotators outperform GPT-3.5 in labeling the Fine-Grained Chinese NER Dataset for the UAV delivery system.

\begin{acks}
This work was supported in part by National Natural Science Foundation of China under Grant 62371411, the Research Grants Council of the Hong Kong SAR under Grant GRF 11217823, InnoHK initiative, the Government of the HKSAR, Laboratory for AI-Powered Financial Technologies, the Meituan Robotics Research Institute.
\end{acks}

\bibliographystyle{ACM-Reference-Format}
\bibliography{ref}

\end{document}